\documentclass[runningheads]{llncs}
\usepackage[T1]{fontenc}
\usepackage{graphicx}
\usepackage{booktabs}
\usepackage{multirow}

\usepackage{amssymb}

\usepackage{mathtools}
\usepackage{soul}
\sethlcolor{yellow}
\usepackage{makecell} 
\usepackage{tabularx}

\newcounter{subfigure}
\renewcommand{\thesubfigure}{\alph{subfigure}}
\makeatletter
\@addtoreset{subfigure}{figure}
\makeatother
\newcommand{\subfigcaption}[1]{%
  \refstepcounter{subfigure}\textbf{(\thesubfigure)}~#1%
}

\usepackage[colorinlistoftodos,prependcaption,textsize=small]{todonotes}

\begin{document}
\title{Sensor to Pixels: Decentralized Swarm Gathering via Image-Based Reinforcement Learning}
\titlerunning{Sensor to Pixels: Swarm Gathering via Image-Based RL}
%

\author{Yigal Koifman $^{\dagger}$\inst{1}\orcidID{0009-0004-2874-6928} \and
Eran Iceland  $^{\dagger}$\inst{1}\orcidID{0000-0002-2126-3526} \and
Erez Koifman\inst{1}\orcidID{0009-0001-4606-3221} \and
Ariel Barel\inst{1}\orcidID{0000-0003-3275-4264} \and
Alfred M. Bruckstein\inst{1}\orcidID{0000-0001-5669-0037}}
%
\authorrunning{Y. Koifman and E. Iceland}
%

\institute{Technion Israel Institute of Technology, Haifa, Israel\\ 
\email{igal.k@cs.technion.ac.il}\and
Technion Israel Institute of Technology, Haifa, Israel \\
\email{eran.iceland@gmail.com}\and
Technion Israel Institute of Technology, Haifa, Israel \\
\email{erez.koifman@cs.technion.ac.il}\and 
Technion Israel Institute of Technology, Haifa, Israel \\
\email{arielba@technion.ac.il}\and
Technion Israel Institute of Technology, Haifa, Israel \\
\email{freddy@cs.technion.ac.il}\\
$^{\dagger}$These authors contributed equally to this work}

\maketitle              

\begin{abstract}
This study highlights the potential of image-based reinforcement learning methods for addressing swarm-related tasks. In multi-agent reinforcement learning, effective policy learning depends on how agents sense, interpret, and process inputs. Traditional approaches often rely on handcrafted feature extraction or raw vector-based representations, which limit the scalability and efficiency of learned policies concerning input order and size. In this work we propose an image-based reinforcement learning method for decentralized control of a multi-agent system, where observations are encoded as structured visual inputs that can be processed by Neural Networks, extracting its spatial features and producing novel decentralized motion control rules. We evaluate our approach on a multi-agent convergence task of agents with limited-range and bearing-only sensing that aim to keep the swarm cohesive during the aggregation. The algorithm’s performance is evaluated against two benchmarks: an analytical solution proposed by Bellaiche and Bruckstein, which ensures convergence but progresses slowly, and \textit{VariAntNet}, a neural network-based framework that converges much faster but shows medium success rates in hard constellations. Our method achieves high convergence, with a pace nearly matching that of \textit{VariAntNet}. In some scenarios, it serves as the only practical alternative.
\end{abstract}
\section{Introduction}
Swarm robotics has gained significant attention due to the potential to solve complex tasks in dynamic environments. 
Coordination of swarm robotics typically follows either centralized or decentralized control paradigms. While centralized methods rely on a global controller with complete system information, they are often limited by scalability and vulnerability to single-point failures. In contrast, swarm robotics control may rely on a decentralized approach characterized by local decision-making, self-organization, and resilience to individual agent failures.

A branch of swarm robotics, known as Ant robotics, focuses on agents that operate with limited sensing, minimal computational resources, and limited or non-existent communication capabilities. This enables scalable and cost-effective real-world applications such as search and rescue, environmental monitoring, and more.
At the same time, they introduce significant challenges for coordination and collective behavior, as agents act based solely on their local sensory input.

Connectivity and convergence play a critical role in swarm missions. Connectivity ensures agents remain linked within the visibility graph, enabling cooperation and coordination. Convergence gathers all agents into a small area, supporting tasks such as rendezvous or resource sharing, and highlights how local rules can generate coherent global behavior.

The primary objective of this work is to explore a Multi-Agent Reinforcement Learning (MARL) based algorithm to generate a decentralized control strategy that enables agents to achieve global objectives. The concept is evaluated on a fundamental task of swarm convergence, while avoiding swarm fragmentation.
In MARL, effective policy learning depends on how agents process and interpret sensory inputs. Traditional approaches often rely on handcrafted feature extraction or raw vector-based representations, which can limit the scalability of learned policies concerning varying sizes and orders of observations.

\begin{figure}[ht]
\begin{center}
\includegraphics[width=8cm]{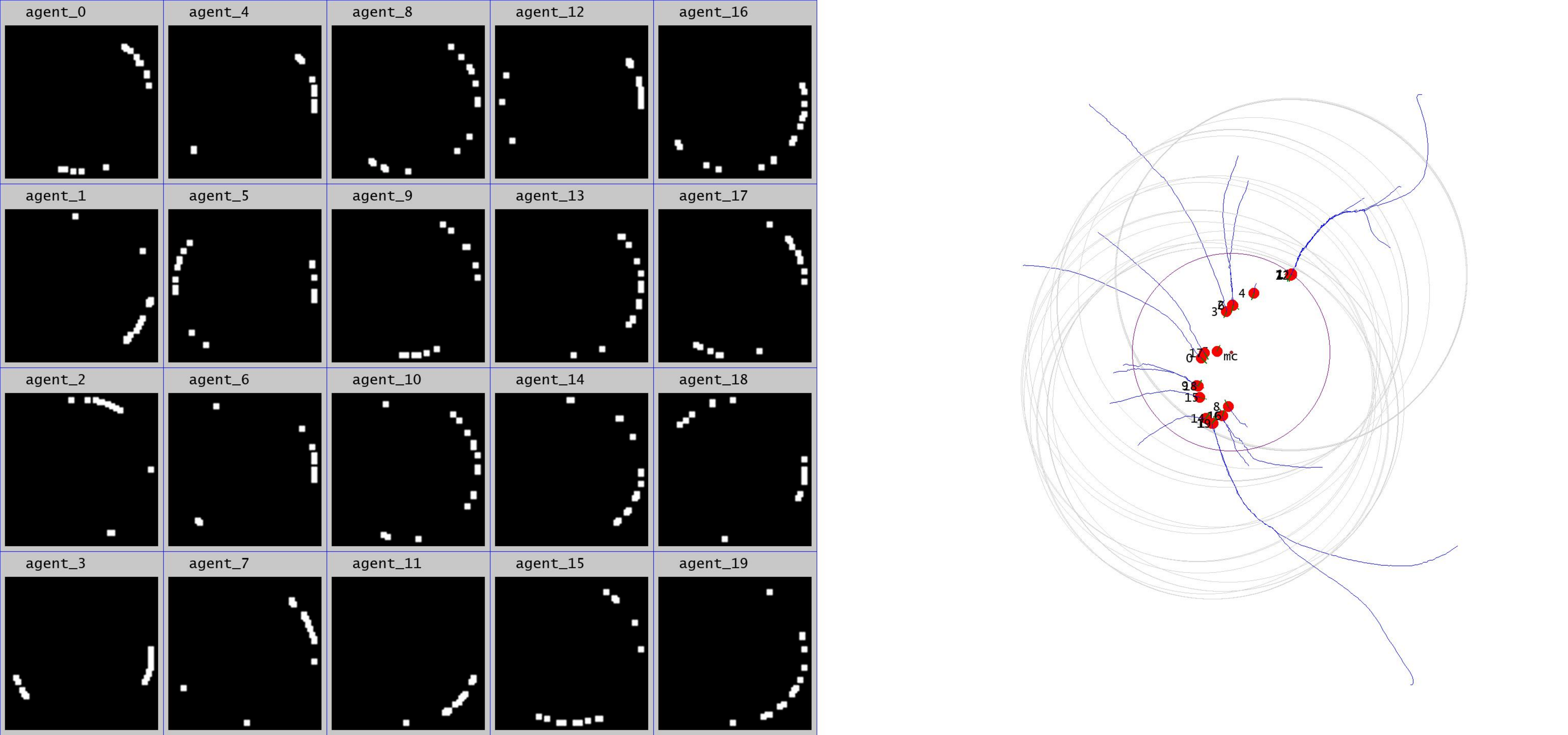}
\caption{Convergence of 20 agents. Left: local observations of a typical agent, where white pixels indicate detected neighbors within a limited sensing range. Right: global view of the swarm’s convergence trace.}
\label{fig: RL snapshot}
\end{center}
\end{figure}

The presented framework implements Centralized Training and Decentralized Execution (CTDE) with an image-based sensing representation, where the agent's observation is encoded as structured visual inputs and processed by a Convolutional Neural Network (CNN), allowing effective extraction of spatial features and enhancing the emergence of control rules as depicted in Fig~\ref{fig: RL snapshot}.

An analytical solution to the gathering problem with finite visibility and bearing-only sensing, i.e., agents can sense only relative direction to their neighbors, was studied by Bellaiche and Bruckstein~\cite{bellaiche2017continuous}, where a local motion rule was proposed to guarantee agents gathering within a finite time. 

We compare the performance of our Reinforcement Learning (RL) policy against the analytical solution and against VariAntNet~\cite{koifman2025variantnetlearningdecentralizedcontrol}, a NN-based framework.
The results demonstrate that the RL policy preserves swarm connectivity even in highly challenging constellations, while achieving a significantly faster convergence rate than the analytical approach, sacrificing guaranteed connectivity throughout the entire process.
Our key contributions:

\begin{enumerate}
    \item \textbf{Image-based bearing representation}: an encoding of bearing-only agent observation as an image, enabling the use of convolutional NN. This approach appears to be novel within the domain of decentralized swarm coordination.
    \item  \textbf{Connectivity-oriented RL control}: an RL policy that strongly promotes swarm connectivity and accelerates convergence compared to the analytical solution, outperforming prior approaches under complex constellations.
    
\end{enumerate}


\section{Related Work}
\label {sec: Related Work}
MARL has become a cornerstone in the study of distributed decision-making, enabling agents to learn cooperative or competitive behaviors in shared environments. Foundational works such as those by Busoniu et al.~\cite{busoniu2008comprehensive} and Zhang et al.~\cite{zhang2019multi} laid the theoretical groundwork for MARL, highlighting its connections to game theory, control, and distributed optimization. More recent surveys~\cite{hernandez2019survey},~\cite{albrecht2023marlbook} have expanded on these foundations, emphasizing the challenges of non-stationarity, credit assignment, and scalability in multi-agent settings.

A widely adopted method in MARL is the CTDE, which allows leveraging global information during training while acting independently at execution time. This approach addresses the instability caused by non-stationary environments and has been formalized in MADDPG by Lowe et al.'s ~\cite{lowe2017multi} and further explored in theoretical treatments like Amato's introduction to CTDE~\cite{amato2024ctde}. The latter has proven especially effective in cooperative tasks, where agents must coordinate without direct communication during deployment.

In the context of swarm robotics, MARL has been applied to tasks such as formation control, rendezvous, and pursuit-evasion. Hüttenrauch et al.~\cite{huettenrauch2019deep} proposed a mean embedding representation for scalable swarm learning, addressing the challenge of representing variable-sized agent groups. These approaches often assume full observability or communication, whereas limited range visibility with bearing-only models, where agents can sense only the relative direction of their neighbors, poses a more constrained sensing paradigm. Recent studies have begun to investigate MARL under limited sensing conditions, such as bearing-only observations. Li et al.~\cite{li2025bearing} proposed a cooperative MARL framework for target pursuit using only bearing measurements, addressing challenges in state estimation and control under minimal sensing. Similarly, Kang et al.~\cite{kangma} introduced the Multi-Agent Masked Auto-Encoder (MA$^2$E), which enables agents to infer global information from partial observations without explicit communication, enhancing decentralized coordination in partially observable environments.

An analytical solution to the gathering problem under the constraints of finite visibility and bearing-only sensing was presented by Bellaiche and Bruckstein \cite{bellaiche2017continuous}.
Their approach requires each agent to identify the smallest circular sector of its visibility range that encompasses all its neighbors. If this sector spans an angle smaller than $\pi$, the agent moves by setting its velocity vector as the sum of two unit vectors directed toward its outermost neighbors. Otherwise, the agent is considered to be surrounded, and it remains stationary.
This is proven to guarantee convergence for any initial configuration and swarm size.

Koifman et al.~\cite{koifman2025variantnetlearningdecentralizedcontrol} propose the VariAntNet, a NN-based approach that addresses the same challenge. It employs a supervised CTDE learning approach to train a stateless neural network using geometric features and a visibility-graph Laplacian-based loss.

While in VariAntNet the control policy is temporally optimized with centralized access and trained to imitate a predefined behavior, our MARL-based approach enables agents to learn decentralized policies through local interaction, optimizing long-term cumulative rewards. This supports richer temporal behavior and greater flexibility, while preserving decentralized execution.
Limited sensing forces the analytical method to adopt cautious rules, resulting in a much slower convergence rate than statistically trained solutions.

\section{Problem Setting}
\label{sec: Problem Setting}
The primary objective of this work is to investigate a MARL framework for representing multi-agent problems. 
As a test case, we examine a fundamental multi-agent gathering task, in which agents must converge within a small area. The agents are defined as having a limited-range and bearing-only sensing, which enables them to detect only the direction of nearby neighbors located in their visibility range.
A key constraint is that the swarm must remain cohesive and prevent fragmentation, which is achieved by following the conservative principle of ``never lose a neighbor''~\cite{barel2019cometogether}, meaning that agents should avoid losing any of their sensed neighbors.

\subsection{Notation and Definitions}
The positions of agents at time $t$ are denoted by \({P}(t) \triangleq \{{p}_i(t)\}_{i=1, 2, 3, \ldots,	N}\) where each agent's position is represented as \({p}_i(t) \triangleq \{(x_i,y_i)^T\}_{i=1, 2, 3, \ldots,N}\).
The visibility range of the agents is denoted as \(V\), and the distance between $p_i$ and $p_j$ is defined as the Euclidian distance: 
    \(d(p_i, p_j) = \parallel{p_j} - {p_i} \parallel \triangleq{{\left((p_j-{p_i})^T ({p_j} -{p_i})\right)}}^{\frac{1}{2}}
    \label{eq: distance_definition}\).
Agent \(i's\) neighbor set, \(\mathcal{N}_i\), is defined as
\(\mathcal{N}_i=\left\{p_j\in P, d(p_i,p_j) \leq V\right.\}\). 
In bearing-only sensing agent \(i\)'s observation of agent \(j\) at time \(t\) is defined as the unit vector,
\(\hat{u}_{ij}\), directed from \(p_i(t)\) to \(p_j(t)\):
\( \hat{u}_{ij}(t)=\frac{p_j(t) - p_i(t)}{\parallel p_j(t) - p_i(t) \parallel}\)


\section{Methodology}
\label{sec: Methodology}



The Sensor to Pixels framework implements a CTDE paradigm within a MARL setting, utilizing an Actor-Critic architecture. During training, centralized data is leveraged, providing access to global environmental information and enabling the computation of global-based rewards based on the swarm’s actions. During execution, each agent functions independently in a decentralized manner, relying solely on its partial local observation.

The problem setting can be modeled as a Decentralized Partially Observable Markov Decision Process (DEC-POMDP)~\cite{bellman1957markovian} which is used to model multi-agent decision-making under uncertainty, where agents must cooperate to achieve a common goal while having partial observations of the environment:
\begin{equation}
\mathcal{M}=\langle \mathcal{S}, \mathcal{N}, \{ \mathcal{O}_i\}, \{ \mathcal{A}_i\}, \mathcal{P}(s' \mid s, \mathbf{a}), \{ \mathcal{R}_i\}, \gamma \rangle, \end{equation}
where a set of \(\mathcal{N}\) agents operate in a shared environment with only partial information of the actual global state, \(\mathcal{S}_t\), thus each agent \(i\) observes only a local limited view of the environment, defined as \(\mathcal{O}_i(t) =  f(\mathcal{S}_t)\).
Therefore, \(\mathcal{S}\) represents the global state, \(\mathcal{N}  = \{1,2, \cdots, N\} \) is a set of agents, \(\mathcal{O}_i\) is the observation of agent $i$, \(\mathcal{A}_i\), the action space of agent i, \(\mathcal{P}(s' \mid s, \mathbf{a})\) is the environment's probability of transitioning from \(s\) to \(s'\) given action a, 
\(\mathcal{R}_i\), the reward function of agent $i$, and \(\gamma\) is the discount factor.

The goal is to optimize the policy function \(\pi\), during training to maximize the expected value of the cumulative discounted reward: \(J(\theta) = \mathbb{E} \left[ \sum_{t=0}^{T} \gamma^t R_t \right]\), where \(\theta\) represents the parameters of the policy network.

\begin{figure}[ht]
\begin{center}
\includegraphics[width=12cm]{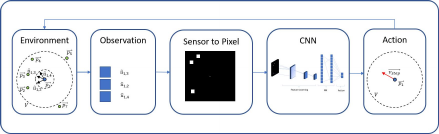}
\caption{Sensor to Pixels framework.}
\label{fig: sensor_to_pixel}
\end{center}
\end{figure}

At each time step, every agent senses its local environment, obtaining a partial observation. This observation is preprocessed and converted into a pixel-based representation. The resulting data is then passed through a CNN-based feature extractor, which encodes the relevant information before it is utilized for policy learning, as illustrated in Fig~\ref{fig: sensor_to_pixel}.
The policy is being updated during training utilizing a reward function that consists of a local and a global reward functions. The local reward function evaluates the agent’s individual actions and states, while the global reward function promotes the behavior of the entire swarm, encouraging coordinated actions.
This design allows agents to learn effective individual behaviors while promoting the global swarm goal to converge to a small area.

\subsection{Observation Space}

\begin{figure}[h!]
    \centering
    \begin{minipage}[t]{0.37\columnwidth}
        \centering
        \includegraphics[width=0.9\columnwidth]{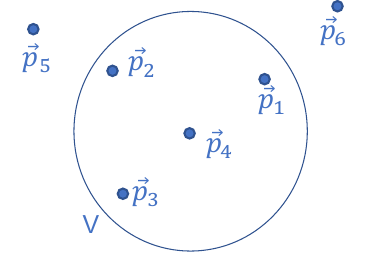}
        \subfigcaption{An environment view around \(p_4\), where \(p_4\) senses the direction of \(p_1,p_2,p_3\).}
        \label{fig:preproc:a}
    \end{minipage}\hfill
    \begin{minipage}[t]{0.25\columnwidth}
        \centering
        \includegraphics[width=0.9\columnwidth]{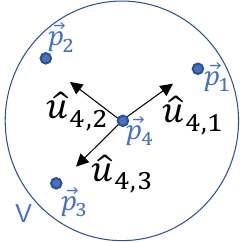}
        \subfigcaption{\(p_4\) observation is \(\hat{u}_{4,1}, \hat{u}_{4,2}, \hat{u}_{4,3}\).}
        \label{fig:preproc:b}
    \end{minipage}\hfill
    \begin{minipage}[t]{0.25\columnwidth}
        \centering
        \includegraphics[width=0.9\columnwidth]{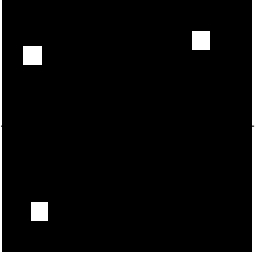}
        \subfigcaption{All \(\hat{u}_{ij}\) are projected onto a binary matrix.}
        \label{fig:preproc:c}
    \end{minipage}

    \caption{Preprocessing from local observation to pixel-grid representation.}
    \label{fig:preprocessing-stage}
\end{figure}

Each agent observes its neighboring agents using its bearing-only, limited-range sensor. As a result, the observation of agent $i$, at time $t$, denoted by \(\mathcal {O}_i(t)\), and defined as a multi-set of unit vectors pointing at its visible neighbors:
\begin{equation}
    O_i(t)=\left\{\hat{u}_{ij}(t) : \forall p_j \in \mathcal{N}_i(t) \right\}
    \label{eq: observation_definition}
\end{equation}
This observation is projected onto a binary matrix, sized $75\times75$ pixels, where agent $i$ is located at the centeral pixel, and where each detected agent in the observation is visualized as $3\times3$ matrix block, positioned according to its relative direction with respect to the agents' local frame of reference as depicted in Fig.~\ref {fig:preprocessing-stage}.


\subsection{Action Space}
In each step, each agent selects an action from a continuous action space, \(\mathcal{A}\). This action is  defined by a direction angle \( \alpha\) and a step size \( \sigma\):
\begin{equation}
\mathcal{A} = \left\{ (\alpha, \sigma) \mid \alpha, \sigma \in \mathbb{R},\ -\pi < \alpha \leq \pi,\ 0 \leq \sigma \leq 1 \right\}
\label{eq: Action_definition}
\end{equation}
The agent’s policy \( \pi(a \mid o) \) maps local observations to actions \( a = (\alpha, \sigma) \), allowing for fine-grained directional control and adaptive motion planning. The angle \( \alpha \) determines the heading, while the scalar \( \sigma \) modulates the movement magnitude. This continuous control setting enables smooth and precise coordination, and promotes maintaining cohesion and accelerating convergence rate in decentralized swarm behaviors.

\subsection{Network Architecture}
\subsubsection{Feature Extraction}
The next stage in the framework is feature extraction, performed by a CNN. The network processes the projected observations, learning to extract spatial patterns and identify embedded features. One of the key advantages of using a CNN is its ability to capture spatial and geometric relationships between observations, regardless of their order, quantity, or orientation.

The architecture is composed of multiple convolutional layers, drawing inspiration from the pioneering work of Mnih et al.~\cite{mnih2015human}, who demonstrated the effectiveness of CNN-based feature extraction in learning to play Atari games at a human-level performance.
The layers and architecture are defined in Table~\ref{tab: CNN pipeline}.

\begin{table}[h]
    \centering
    \caption{The CNN Pipeline.}
    \renewcommand{\arraystretch}{1}
    \resizebox{0.8\linewidth}{!}{
    \begin{tabular}{c c c c c c}
        \hline
        \textbf{Layer} & \textbf{  In Channel  } & \textbf{  Out Channel  } & \textbf{  Kernel Size  } & \textbf{  Strides  } & \textbf{  Activation  } \\ 
        \hline
        Layer 1 & 1 & 32 & 8,8 & 4 & ReLU \\
        Layer 2 & 32 & 64 & 4,4 & 2 & ReLU \\
        Layer 3 & 64 & 64 & 3,3 & 1 & ReLU \\
        \hline
    \end{tabular}}    
    \label{tab: CNN pipeline}
\end{table}

The output from the final convolutional layer is first flattened into a vector of 1600 neurons, which is then processed by a multilayer perceptron (MLP) to reduce its dimensionality to a 512-feature representation.

\subsubsection{Actor-Critic Network}
\begin{figure}[!h]
\begin{center}
\includegraphics[width=8cm]{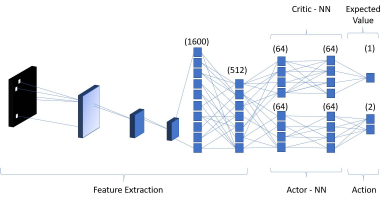}
\caption{Actor Critic NN Architecture.}
\label{fig: Actor Critic NN Architecture}
\end{center}
\end{figure}

While the actor and critic networks share a common feature extraction module, their subsequent architectures are independent. The actor network deterministically outputs the specific action to be taken in the current state, directly mapping the learned policy to an action vector. The critic network outputs a scalar value estimating the expected value function for the current (state, action) pair, which is used to evaluate and improve the policy as depicted in Fig~\ref{fig: Actor Critic NN Architecture}.
\subsection{Reward Function}
The objective is to converge the swarm to a compact area while maintaining cohesion and preventing fragmentation. This collective behavior emerges from the learned policy, which is shaped by the reward function.  The agent’s reward function comprises two components:
\begin{itemize}
    \item \textbf{Local Reward function}: \(R_{local}^i\), is designed to promote swarm cohesion similar to the ``never lose a neighbor” principle described in~\cite{manor2019local}: an agent starting in a connected state should maintain visibility with its set of accumulated neighbors. Therefore, a neighbor-loss penalty, \(P_{ln}\), is applied to the local reward for each single neighbor lost wherever the loss occurs.
    \(R_{ln}^i(t)\) represents the total penalty accumulated at time step \(t\) due to the loss of neighbors.
     \[R_{ln}^i(t) = (|\mathcal{N}_i(t)| - |\mathcal{N}_i(t+1)| )\cdot P_{ln}, \quad for \hspace{0.2 cm} |\mathcal{N}_i(t+1)| < |\mathcal{N}_i(t)|\]
    To accelerate the swarm convergence and reduce the episode steps, an additional penalty, \(P_{acc}\), is applied for every step performed by each agent. The local reward is therefore given by 
    \(R_{local}^i(t)  = R_{ln}^i(t) + P_{acc}\).
    \item \textbf{Global Reward}: \(R_{global}\), is designed to promote swarm convergence by minimizing the bounding radius, \(D_{\text{global}}(t) \), of a circle centered at the mean position of the agents. 
    
    The bounding radius is defined as:
    \( D_{\text{global}}(t) = \max_{i \in \mathcal{N}} || p^i(t) - C_{\text{swarm}}(t) ||,\)
    where the mean position is given by
    \(C_{\text{swarm}}(t) = \frac{1}{N} \sum_{i=1}^{N} p_i(t)\) and the global reward is computed as
    \( R_{\text{global}}(t) = (D_{\text{global}}(t) - D_{\text{global}}(t+1))\cdot C_g,\)
where $C_g$ is a normalization constant.
\end{itemize}
The reward, \(\mathcal{R}^i \), for agent \(i\), is defined as 
\(R^i(s, a^i, \mathbf{a}) = R^i_{\text{local}}(s, a^i) + R_{\text{global}}(s, \mathbf{a}).\)


\section{Results}
\label{sec: Experimental Results}
The framework is evaluated through a simulation-based approach. First, we train the model using a structured learning process to optimize its performance. Second, the trained model is evaluated across a diverse set of scenarios to assess its adaptability and effectiveness. Last, its performance is compared against the analytical algorithm~\cite{bellaiche2017continuous} and VariAntNet by measuring two key metrics: the percentage of scenarios that were successfully gathered, and the convergence time.
The policy is trained and evaluated using the PettingZoo library~\cite{terry2021pettingzoo} 
for multi-agent environment simulation, in combination with Stable-Baselines3~\cite{stable-baselines3}.

\label{sec: rl training}
Our dataset contains connected constellations of randomly positioned agents with varying sizes, visibility ranges, and difficulty levels. The difficulty is determined by the initial spatial distribution of agents and quantified using the effective visibility range,\(V_{\text{eff}}=V\times VR,\)
where $0 < VR \leq 1$ is the visibility ratio. During generation, agents are placed sequentially, starting from a random initial position. Each new agent is repeatedly positioned until it lies within $V_{\text{eff}}$ of at least one previously placed agent, ensuring connectivity. Increasing $VR$ raises the difficulty, as the risk of disconnection grows. We define constellations with $V_{\text{eff}} = 0.75$ as `Challenging' and those with $V_{\text{eff}} = 1$ as `Marginal'.

\subsection{Training Parameters} 

We employ the multi-agent variant of the Proximal Policy Optimization (PPO) algorithm~\cite{schulman2017ppo}, which has demonstrated strong performance in single-agent RL tasks. Recent studies~\cite{yu2021surprising} have shown that PPO can be effectively extended to cooperative multi-agent environments, making it a suitable choice for our task.

The visibility range is set to 50 and the step size to 0.5. The training curriculum of the model consists of a learning phase of 150M steps on 10-agent constellations, followed by another 150M steps on 20-agent constellations. Episode cut-offs were 1500 and 3000 steps per agent, respectively.
Our algorithm was trained with a learning rate of 0.00002, a discount factor $\gamma = 0.95$, a batch size of 2048, and 8 parallel environments.
The reward function parameters  are: \(P_{ln}= -0.5\),  \(P_{acc}= -0.01\), and \(C_g= 0.1\).

\subsection{Training Evaluation}
\begin{figure}[htbp]
    \centering
    \begin{minipage}[b]{0.49\textwidth}
        \centering
        \includegraphics[width=\textwidth]{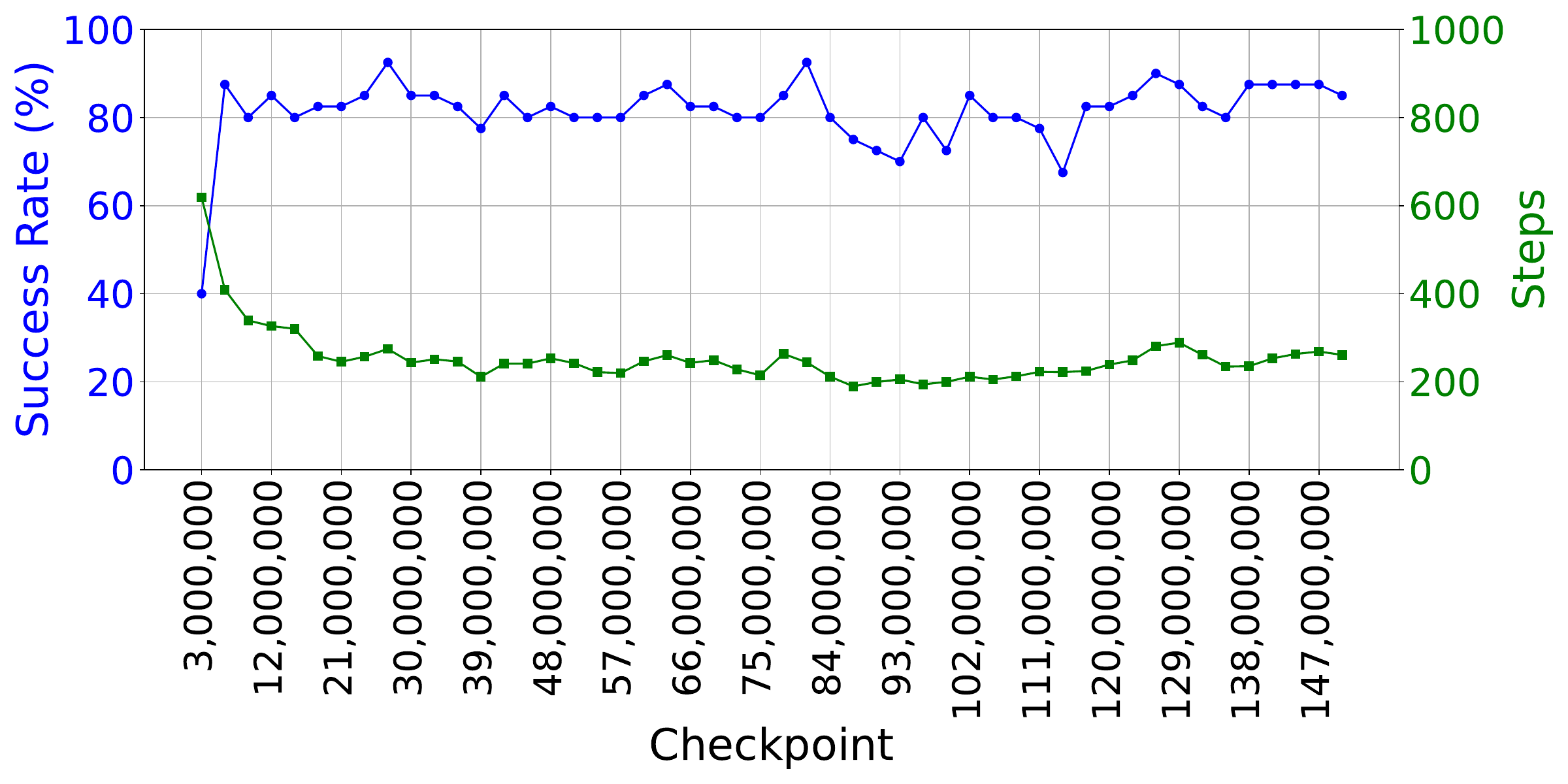}
        \subfigcaption{RL10 training process evaluation.}
        \label{fig: RL10}
    \end{minipage}
    \begin{minipage}[b]{0.49\textwidth}
        \centering
        \includegraphics[width=\textwidth]{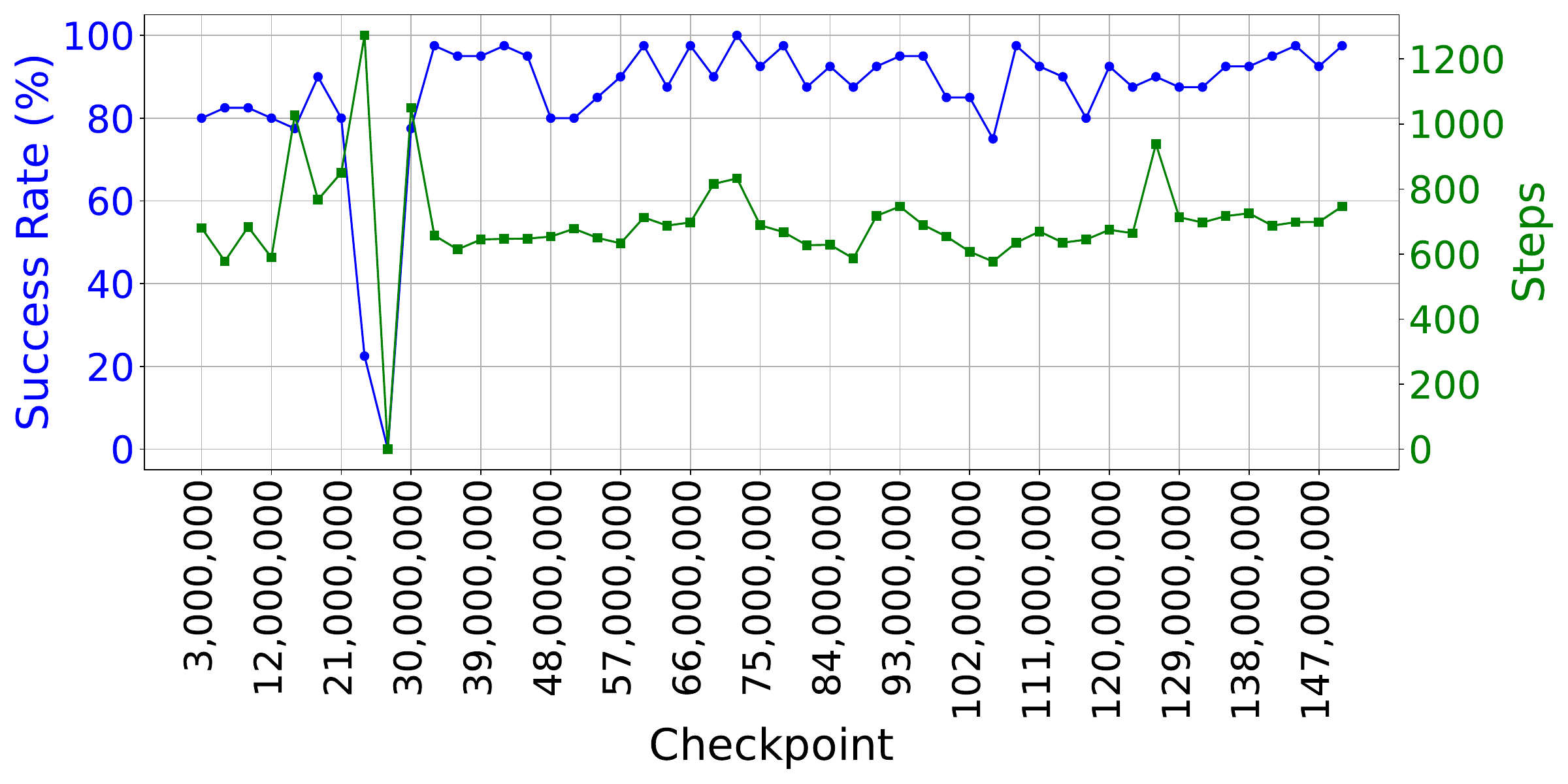}
        \subfigcaption{RL20 training process evaluation.}
        \label{fig: RL20}
    \end{minipage}

    \caption{Training process evaluation of 10 and 20 agents after a convergence to a stable policy. From each training session, a single checkpoint model was selected for further evaluation.}
    \label{fig: checkpoints}
\end{figure}

To assess the training process, checkpoints were stored at intervals of 3M steps. For both the 10 and the 20 agents settings, evaluation was performed on a fixed set of 40 constellations at each checkpoint. The results are summarized in Fig~\ref{fig: checkpoints} and show that the policy rapidly converged to a strong solution during the initial training phase. Beyond this point, performance displayed noticeable fluctuations over the remaining course of training.
The selected checkpoint is the one that best combines a high convergence rate with fast convergence. That is, for RL10, the checkpoint model 81M, and for RL20, the checkpoint model 42M.



\subsection{Execution Results}
The model is evaluated on `Challenging' and `Marginal' initial constellations \( VR=(0.75,1)\). 
The results are compared against the analytical solution and the VariAntNet methods as summarized in Table~\ref{tab: model_comparison}.
The results indicate that agents trained with image-based sensing representations achieve faster and more reliable convergence in `Marginal' initial configuration. Under 
`Challenging' conditions, RL20 achieved 100\% connectivity in 506 steps for a swarm of 20 agents, surpassing VariAntNet (94.9\%, 322 steps) in connectivity preservation, albeit at the expense of longer convergence time, while still converging faster than the analytical baseline (100\%, 702 steps).
We define that a result is acceptable if connectivity exceeded 90\%.
In marginal conditions with 30 agents, VariAntNet exhibited notable performance degradation, while the analytical model demonstrated the slowest convergence, and RL20 maintained superior performance with only moderate performance loss.
RL20 is the only viable RL variant in Marginal settings, outperforming the analytical baseline by 2–28\%. This is attributed to the RL framework’s ability to assign negative rewards 
for any disconnection between agents and positive rewards for swarm convergence. Since RL considers not only the next step reward but 
also future rewards, the model can learn optimal strategies over time. 
In Challenging settings, the weighted VariAntNet outperforms RL10 by 15\%.

\begin{table*}[t]
\centering
\renewcommand{\arraystretch}{1.1}

\newcommand{\thSteps}{\textbf{Steps}}
\newcommand{\thConn}{\makecell[t]{\textbf{Conn}\\[-0.4ex](\%)}}

\setlength{\tabcolsep}{1.5pt}
\caption{
Comparison of the analytical model by Bellaiche~\cite{bellaiche2017continuous}, VariAntNet (Max) and VariAntNet (Max, Weighted 1,10), and MARL methods RL20, RL10 in 1000 scenarios of Challenging and Marginal initial configurations. Reported are average convergence times and the percentage of agents that converged while preserving connectivity. Highlighted results indicate the best performance under the acceptable criteria.}
\resizebox{\textwidth}{!}{%
\begin{tabular}{l|cc|cc|cc|cc|cc|cc}
\hline
\multirow{3}{*}{\textbf{Model}} &
\multicolumn{6}{c|}{\textbf{Challenging}} & \multicolumn{6}{c}{\textbf{Marginal}} \\
& \multicolumn{2}{c|}{\textbf{10 Agents}} & \multicolumn{2}{c|}{\textbf{20 Agents}} & \multicolumn{2}{c|}{\textbf{30 Agents}} 
& \multicolumn{2}{c|}{\textbf{10 Agents}} & \multicolumn{2}{c|}{\textbf{20 Agents}} & \multicolumn{2}{c}{\textbf{30 Agents}} \\

& \thSteps & \thConn & \thSteps & \thConn & \thSteps & \thConn 
& \thSteps & \thConn & \thSteps & \thConn & \thSteps & \thConn \\

\hline
Analytical
& 271 & 100.0 & 702 & 100.0 & 1235 & 100.0
& 362 & 100.0 & 950 & 100.0 & 1629 & 100.0 \\
VariAntNet(Max)
& 165 & 99.5 & 346 & 95.2 & 554 & 91.6
& 217 & 84.5 & 451 & 58.4 & 711 & 45.8 \\
VariAntNet(Max, W)
& \textbf{159} & \textbf{98.8} & \textbf{322} & \textbf{94.9} & \textbf{506} & \textbf{90.6}
& 209 & 82.6 & 418 & 53.8 & 646 & 41.3 \\
RL20
& 241 & 100.0 & 506 & 100.0 & 776 & 99.8
& \textbf{354} & \textbf{97.0} & \textbf{753} & \textbf{93.6} & \textbf{1173} & \textbf{91.6} \\
RL10
& 186 & 99.7 & 383 & 95.4 & 565 & 89.2
& 260 & 82.9 & 543 & 61.7 & 797 & 40.6 \\
\hline
\end{tabular}}
\label{tab: model_comparison}
\end{table*}

An interesting phenomenon occurs when the training transitions from 10 to 20 agents; the model becomes more conservative, decreasing the number of disconnections at the expense of the convergence rate.  This results from the fact that 
as the number of agents increases, the probability of having ``leaf" agents in the initial constellation (those with only one neighbor) also increases. Disconnecting the sole edge linking these agents results in a graph disconnection. Additionally, the probability of having a bridge (cut edge) that, when disconnected, splits the swarm into segments, also increases. Therefore, as the number of agents in the training phase increases, the model learns to behave more conservatively.

\section{Ablation Study} 
\label{sec: Ablation Study}

The ablation study investigates the impact of the local reward, that is, the neighbor loss penalty factor \(P_{\text{ln}}\), on the swarm behavior. We evaluated two additional values \(P_{\text{ln}}= 0 \text{ and }P_{\text{ln}}=-1\), in addition to the baseline value of \(P_{\text{ln}}=-0.5\). For each configuration, the model was trained for 150M steps.

For \textbf{\(P_{\text{ln}} = -1\)}, the learned policy resulted in all agents remaining stationary. The penalty for losing a neighbor was disproportionately high, leading agents to be very conservative and to avoid any movement that might risk disconnection.

For \textbf{\(P_{\text{ln}} = 0\)}, here, the opposite behavior was observed. Agents were eager to converge and ignored the consequences of losing connections. In many instances, the swarm remained connected despite the loss of individual edges. This led to rapid convergence, which came at the cost of reduced overall convergence quality.

Table~\ref{tab:ablation_study} compares the results of this configuration with the baseline penalty of \(P_{\text{ln}}=-0.5\) (RL10 from Table~\ref{tab: model_comparison}). For 10 agents constellation with \(V_{\text{eff}} = 0.75\), the system achieved a reasonable convergence rate with fast convergence. However, for 20 agents constellation or when \(V_{\text{eff}} = 1\), the convergence rates dropped significantly.
\begin{table*}[h!]
\renewcommand{\arraystretch}{1.1} 
\caption{Ablation study results for different neighbor loss penalty values $P_{\text{ln}}$.}
\centering
\resizebox{0.7\textwidth}{!}{%
\begin{tabular}{l @{\hspace{5pt}} c c @{\hspace{5pt}} c c @{\hspace{5pt}} c c@{\hspace{5pt}} c c}
\toprule

\multirow{3}{*}{}
  & \multicolumn{4}{c}{\textbf{10 Agents}}
  & \multicolumn{4}{c}{\textbf{20 Agents}} \\
 
\multirow{2}{*}{$V_{\text{ref}}$} &
\multicolumn{2}{c}{$P_{\text{ln}} = 0$} & 
\multicolumn{2}{c}{$P_{\text{ln}} = -0.5$} & 
\multicolumn{2}{c}{$P_{\text{ln}} = 0$} & 
\multicolumn{2}{c}{$P_{\text{ln}} = -0.5$} \\ 
\cmidrule(lr){2-3} \cmidrule(lr){4-5} \cmidrule(lr){6-7} \cmidrule(lr){8-9}
& Steps & Conv.~(\%) & Steps & Conv.~(\%) & Steps & Conv.~(\%) & Steps & Conv.~(\%) \\ 
\midrule
0.75 & 147 & 92.1 & 186 & 99.7 & 245 & 59.8 & 383 & 95.4\\ 
1.00 & 193 & 55.2 & 260 & 82.9 & 313 & 11.6 & 543 & 61.7\\ 
\bottomrule
\end{tabular}}
\label{tab:ablation_study}
\end{table*}

\section{Limitations}
\label{sec: Limitations}

A notable limitation of RL based approaches is their long training time and cherry-picking process to identify better learning parameters, which often requires extensive computational resources and careful hyperparameter tuning. Even after prolonged training, these methods can not guarantee convergence to a stable or optimal policy, as performance may fluctuate across runs or fail to generalize to unseen scenarios.


    


\section{Conclusion}
\label{sec: Conclusion}

In this work, we introduce the Sensor-to-Pixels framework, a MARL-based approach that integrates CTDE and CNN-based perception for decentralized swarm coordination. The results show that the framework significantly outperforms 
known models achieving faster convergence and stronger cohesion across varying swarm sizes. 
These findings highlight the potential of RL-based methods to address fundamental challenges in swarm robotics, particularly in time-critical scenarios such as disaster response. 
Looking ahead, future research will explore more complex multi-agent behaviors and environments. An additional research path lies in developing an adaptive decision framework that dynamically selects the most suitable algorithm during inference, balancing analytical guarantees with learning-based efficiency.

\bibliographystyle{splncs04}
\bibliography{bib}
%




\end{document}